# Influence of the learning method in the performance of feedforward neural networks when the activity of neurons is modified

M. Konomi and G. M. Sacha.

Departamento de Ingeniería Informática. Escuela Politécnica Superior. Universidad Autónoma de Madrid. Campus de Cantoblanco. 28049 Madrid. Spain

Abstract:

A method that allows us to give a different treatment to any neuron inside feedforward neural networks is presented. The algorithm has been implemented with two very different learning methods: a standard Back-propagation (BP) procedure and an evolutionary algorithm. First, we have demonstrated that the EA training method converges faster and gives more accurate results than BP. Then we have made a full analysis of the effects of turning off different combinations of neurons after the training phase.

We demonstrate that EA is much more robust than BP for all the cases under study. Even in the case when two hidden neurons are lost, EA training is still able to give good average results. This difference implies that we must be very careful when pruning or redundancy effects are being studied since the network performance when losing neurons strongly depends on the training method. Moreover, the influence of the individual inputs will also depend on the training algorithm. Since EA keeps a good classification performance when units are lost, this method could be a good way to simulate biological learning systems since they must be robust against deficient neuron performance. Although biological systems are much more complex than the simulations shown in this article, we propose that a smart training strategy such as the one shown here could be considered as a first protection against the losing of a certain number of neurons.

Highlights:
- The ability to change the individual performance of neurons have been added to feedforwad neural networks
- Feedforward Neural Networks improves the classification results when trained by evolutionary algorithms
- The use of evolutionary algorithms keeps a good performance of the networks when individual units stop working properly
- Robustness against death of neurons can be simulated by feedforward neural networks trained by evolutionary algorithms

# 1. Introduction

Over the last decades, solving classification problems with artificial neural networks such as Feedforward Neural Networks (FFNN) is considered a promising and useful strategy [1] due to their high capability to classify real world complex problems. In classification problems, the network has to be trained (i.e. generate a set of weights that solves the problem properly) with a set of data to correctly classify their desired outputs. Once trained, the network is ready to classify new data [2]. Since training means to optimise the weights of the network, we should decide which optimization algorithms better perform the task. At present, there are a lot of different strategies used to train a network. One of the most widely used is the backpropagation algorithm (BP) [3,4]. However, regardless of its popularity, BP has several disadvantages. First, BP converges very slowly when the network's architecture starts being big and complex. Another well-known undesired effect is that BP can easily fall into local minima [5,6]. It is also very dependent of several parameters. For example, to make BP work properly, learning factor and momentum parameters must be chosen carefully. It has been established that, once working, any slightly change of them can disturb the networks accuracy [7]. Finally, BP strongly depends on the training set presentation: to achieve the optimal set of weights, BP depends on the sequence of training cases (i.e. the same cases in a different order may train the network better or worse). It was pointed out by Curry and Morgan [8] that gradient techniques might not always give the best and fastest way to train an FFNN.

In this context, evolutionary algorithms (EAs) appear as a promising answer to the necessary improvement in the FFNN learning process. EAs are global search techniques adopting the principle of natural biological evolution and/or the social behaviour of species. One of the main advantages of EAs is their ability to escape from local minima [9] since, unlike BP, they start with a wide population of solutions. There are already many studies that show how EAs give accurate and promising results [10,11]. For example, in [1, 12] we can see comparatives between BP and genetic algorithm learning (for both real and binary coded), as well as a method by Daniel Rivero et al.[13] to use EA to optimise the network architecture.

In this article, we develop two different training methods for FFNN that are based in BP and EA as well. In both cases, the network units can work independently and their behaviour can be individually modified in any phase along the network optimisation. Using this characteristic, we have developed a comparative study between both training methods, where we analyse the effect of losing different kinds and number of individual units. This study can be useful for pruning or redundancy studies. However, it is also a way to simulate the death or wrong performance of biological neurons. As we will show, EA training will have a better performance in all the cases under study. This result implies that there may be intrinsic factors along the learning process that prevent the memory systems against the losing of neurons, independently of any other biological mechanism.

## 2 Technologies

The different technologies used in this project are described in this section. We will start explaining the main common characteristic of the FFNN. Then, we will go through the basics of the two training techniques with the main modifications specifically developed for the comparative analysis.

### 2.1 Characteristics of the network

We will use a particular case of multilayer FFNN. In general, this network is formed by several neurons or units, which are organized in layers. Every layer generates the input for the next one if any. When a hidden layer is used at least, FFNN can classify sets that are not linearly divisible. From the most widely used FFNN, we have added context-independent functionality to the individual neurons. Now, every neuron can be configured to modify its behaviour in the following ways:

-**Neuron Shutdown.** A specific neuron can be shut down (i.e. turning its output to 0). The effects of turning off input or hidden neurons are shown in figure 1. As we can see, a certain number of weights stop working (or becoming useless since they end in a non-working neuron). Instead of turning off a neuron completely, we can also reduce its performance by a certain percentage, emulating a partial wrong behaviour.

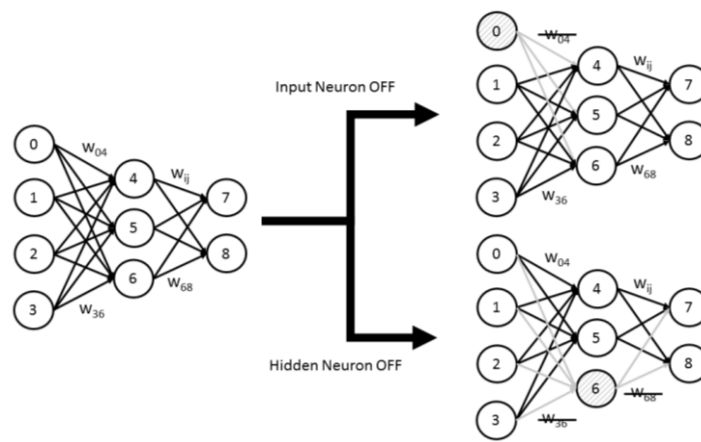

Figure 1: Scheme of the feedforward neural network and the effects on the network performance when an input or hidden layer is turned off. Grey arrows represent the weights that stop having any influence in the final result.

- **Possibility to use a different activation function.** For the present study, we have just implemented the sigmoid function.
- **Sigmoid function modification.** When setting up a neuron with a sigmoid function, we can modify its sigmoid function by changing the parameters α, β and γ, as we show in the following equation:

$$s(x) = \frac{\beta}{\gamma + e^{-x\alpha}} \qquad (1)$$

- **Managing the learning factor.** Our FFNN allows the user to change the learning factor of a particular neuron. The learning factor had an effect on the speed of the learning process. This gives us the opportunity to make a neuron learn faster or slower than others.

## 2.2 Evolutionary Algorithm (EA) learning process

EAs are essentially search functions, i.e. EAs search for the fittest solution of a problem [14]. To decide whether a solution is good or not, the algorithm bases its decision on a pre-set criteria, where the best solution will be the one that better satisfies this criteria. The searching strategy is inspired by the evolutionary process at cellular level. First, EAs generate a set of random solutions we call Chromosomes. Then, they make chromosomes evolve by means of crossing, mutating and selecting the finest next chromosome generations, created from the previous ones. In the evolutionary algorithm that is the origin of our method [15], a chromosome is represented by an array of bits. The process to obtain a solution is divided into the following stages:

1- **Generation of a random set of chromosomes (Initialization).** In this phase the initial set of chromosomes is created. The randomness is needed in order to have the wider variety of elements, which is needed to avoid local minimum and premature wrong convergence of the algorithm

2- **Evaluation of the population.** In this stage, once we have a set of individuals, all of them are evaluated according to the previously selected criteria. This evaluation is done by the fit function. Indeed, depending on the problem, this function can be very different. Since the fit function drives the search of the algorithm, it is extremely important.

3- **Selection of the finest.** Once the population has been evaluated, the best ones are selected to create a new generation of chromosomes that will inherit their characteristics. There are several selection methods, but all of them mainly follow the same principle: better solutions have better probabilities of being selected. First, a selection probability is calculated with the following equation.

$$p(C_i) = \frac{f(C_i)}{\sum_{j=1}^{N} f(C_j)} \qquad (2)$$

where P is the selection probability of the chromosome $C_i$ and the function f(Ci) is the fit function of a chromosome. When all chromosomes have their respective P(Ci) calculated, a selection procedure called stochastic sampling with replacement [155] is used. All chromosomes are mapped on a roulette wheel (see figure 2). The roulette is proportionally divided between chromosomes according to their probabilities. By spinning the roulette, the selected chromosomes are copied for the new population. Because of this mechanism, chromosomes with high probability might be copied several times.

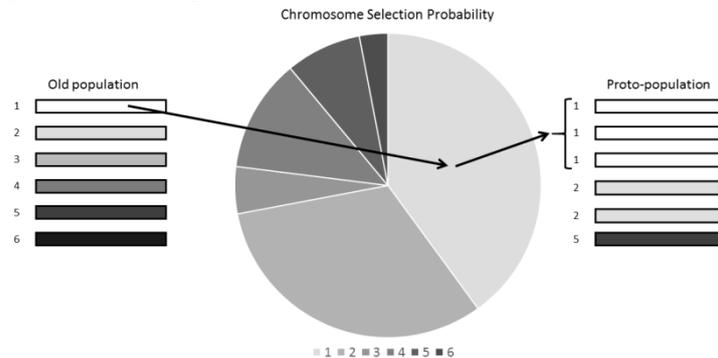

Figure 2: Roulette wheel method mapping the probabilities of the old population (left) and selecting the candidates for the new generation (proto-population on the right side).

Once we have selected enough chromosomes to have a new population, the effect of the selection process produces that the individuals with the best fit will probably be repeated in the new population, taking the place of the worst ones.

4- **Individual Crossing.** Crossing emulates genetic recombination that takes place in meiotic cell division. In that process, two homologous chromosomes exchange pieces of genetic material, turning themselves into different ones. The crossing process is done by selecting an initial crossing point along the chromosome and swapping the following bits with their counterpart. Alternatively, selecting an initial and final crossing point and swapping only the bits between them is another way of individual crossing.

5- **Individual Mutation.** At this point, some randomly chosen chromosomes are mutated by changing the value of a bit from 1 to 0 or vice versa. The reason of this step is exploring new possible solutions that were impossible to reach with no more than the information from the parents. Like in natural selection, mutation can be beneficial for the chromosome or harmful. Since it is likely to be harmful, only a few chromosomes are mutated.

6- **Repetition.** At this point, our new population is ready to go back to phase 2. This loop will continue until we get a solution good enough to solve the problem, or a certain number of generations are reached. Both conditions must be checked between phases 2 and 3.

Now, we will show how we have adapted this general algorithm for the specific objectives of this article. First, we need to represent the solution in a way we can apply mutation and crossing in the way we have shown before. In other words, we need to represent the solution of the problem as a chromosome. To use EA in the training of FFNN, our chromosome will be an array of weights, which correspond to a set of real numbers. Since the original EA was designed with arrays of bits, several modifications are needed to make it work with real numbers.

If we want to keep the classic EA unmodified, we could think, as a first approximation, a way to transform real numbers to a bit based representation (binary coded EA). Although initially this solves the problem, it also limits the accuracy of the algorithm because we transform a continuous representation to a discrete one. Since this first approximation lacks the accuracy, we have chosen to adapt the algorithm to a Real Coded Genetic Algorithms (RCGAs) [16, 17] in the following way:

1- **Chromosome representation.** As we have already stated, chromosomes will become arrays of real numbers instead bits. For this reason, we have to change the full structure of the algorithm.

2- **Evaluation of the population.** The original EA does not specify any evaluation function (fit function) which is problem-dependant. In our solution the fittest chromosomes are the ones that have the lower value in the fit function. For the training of FFNNs, the value is calculated by adding all the least-square errors between the output of the system and the desired output as shown in the following formula.

$$fit(x) = \sum_i (net(i,x) - t_i)^2 \qquad (3)$$

where **x** is a chromosome, **i** represents a training case, **net(i, x)** represents the output of the FFNN for the case **i** and the set of weights of the chromosome **x** and **t** is the desired output for the case **i**. Once the fit of the population is calculated, it is normalised and inverted (i.e. lowest fit function gets the highest probability) for the selection.

3- **Selection of the finest.** Because the evaluation always gives a real value (both in our version of EA and the original) there is no need to change the selection mechanism. It works as previously explained.

4- **Crossing.** This phase of the algorithm is heavily affected by the change of individual representation. If we made it the way it is usually done in the original algorithm, the resulting individuals would be too different from the originals [18] thus making convergence difficult. Instead we calculate the crossing as shown in the following equations:

$$\begin{cases} I'_1 = \alpha \cdot W_1 + (1-\alpha) \cdot W_2 \\ I'_2 = \beta \cdot W_1 + (1-\beta) \cdot W_2 \end{cases} \qquad (4)$$

where $I_1$ and $I_2$, with their respective arrays of weights $W_1$ and $W_2$, are crossed to form $I'_1$ and $I'_2$ using two random numbers α and β that within the range [0-1]. In this way, both new individuals will have a part of their progenitors.

5- **Mutation.** This step has to be modified as well. In the original EA, mutation simply switched a bit to its other possible state, 1 or 0. This cannot be done with real numbers, there is no other state. Following the philosophy of mutation, what we have done is to add or subtract a small number to the weight chosen for mutation. This increment is chosen randomly inside a range defined by a constant [188].

$$\Delta x = \alpha \cdot cte \qquad (5)$$

Being α a random number between -1 and 1, this will make the selected weight **x** to vary from **x-cte** to **x+cte**.

## 2.3 Back Propagation learning strategy

BP is one of the most popular algorithms to train and make the FFNN learn [7]. Since this method is well-known and extensively explained in the bibliography, we are going to give here only the modifications that we have done to the standard BP strategy.

In out method, we have used a sigmoid function S as the transfer function of the network:

$$s(x) = \frac{1}{1+e^{-x}} \qquad (6)$$

where x is the sum of the weights that acts as inputs to the unit. This function is commonly used because its derivative can be written as a function of the sigmoid function itself:

$$\frac{ds(x)}{dx} = s(x)(1-s(x)) \qquad (7)$$

To make the neurons work with a specific and unique behaviour, we have applied the modifications shown in equation 1 to the sigmoid function. The algorithm is able to change the behaviour of individual units both in the training and validation phases. The momentum factor [7] can be also switch off at any time in both phases.

## 3. Results

We have studied 5 different sets [19]: Breast Cancer Wisconsin (Original) (CANCER); Credit Approval (CARD); Pima Indians Diabetes (PIMA); Horse Colic (HORSE) and Connectionist Bench (Sonar, Mines vs. Rocks) (SONAR). In all of them we have applied both BP and EA training. In both cases we have execute the simulation 20 times where the training phase started with different randomly selected initializations. The parameters for the BP training are the following: at least 20000 learning iterations and learning factor 0.1. For the EA training we have used 1000-17000 generations, mutation factor = 0.4, mutation and crossover probability = 0.8 and population size = 10. We have chosen these values after different simulations with a wide variety of values for all the parameters. The final decision is the one that guarantees a good performance for all the sets under study. In table 1 we show the FFNN geometry selected for every set. In figure 3 we show an example of the learning rate for three representative sets (CANCER, CARD and PIMA) for both BP and EA training.

| Set | Inputs | Hidden | Outputs | Class 1 | Class 2 | Class 3 |
|---|---|---|---|---|---|---|
| CANCER | 9 | 8 | 2 | 458/65.52% | 241/34.48% | |
| CARD | 51 | 6 | 2 | 307/44.49% | 383/55.51% | |
| PIMA | 8 | 6 | 2 | 268/34.76% | 503/65.24% | |
| HORSE | 58 | 12 | 3 | 224/61.54% | 88/24.18% | 52/14.29% |
| SONAR | 60 | 12 | 2 | 152/48.72% | 160/51.28% | |

Table 1: Geometry used in the simulations for the different sets. It is also shown the number of cases that correspond to the different classes in the set. These numbers are given in absolute value and percentage.

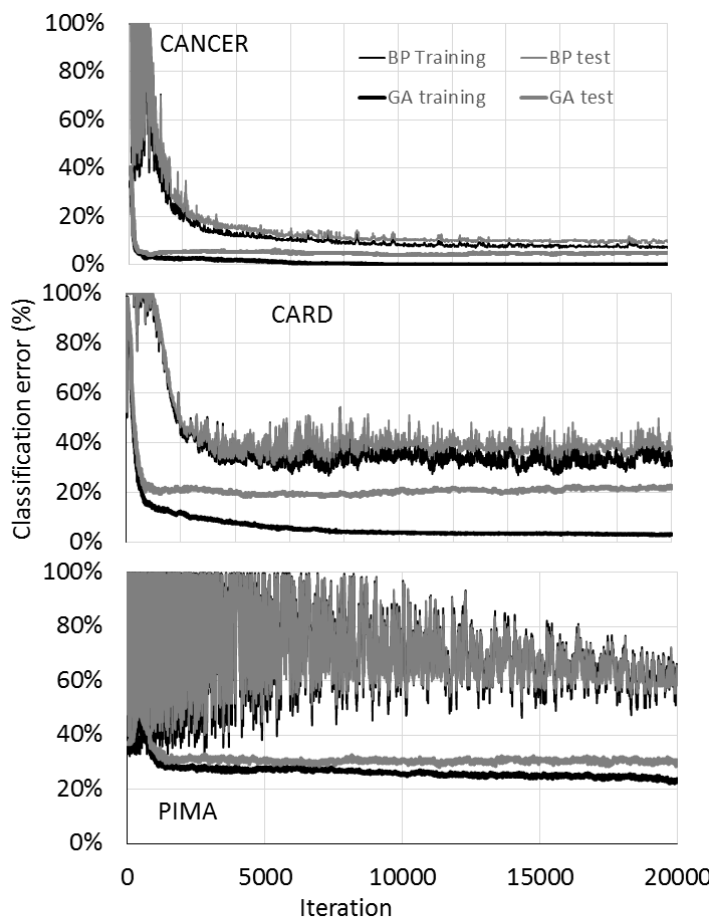

Figure 3: Training evolution for the CANCER, CARD and PIMA sets for both Genetic Algorithm and Backpropagation methods. The error is shown as the percentage of cases that are not correctly classified.

As we can see, the EA based training converges much faster and has a better performance for all the cases under study. Similar results are found for the HORSE and SONAR sets and are not explicitly shown in the figure. In average, EA needs 2.29 times less iterations than BP. In the CARD case, we have found a small overtraining effect in EA training that is not distinguishable for BP. Since this effect is very small and the training set is clearly improved, we have decided to use the values for 20000 iterations in all the figures. As we can see in figure 3, this value is a very good choice for any other set and training method since overtraining is not present. Only in the PIMA set we have used a higher value (200000) since the BP method did not converge well at 20000 iterations.

Let us study now the effect of losing individual neurons or units. To do it, we are going to use the ability of our algorithm to configure different performances of the individual neurons. First, we are going to switch off a single neuron in the hidden or input layer by setting β=0 in the sigmoid function. It is worth noting that this change is only applied after the training phase, i.e. the network has been trained with a full and correct performance of every single neuron in the system. After that, we are going to turn off sets of two hidden neurons. Since the neurons in the hidden layer are initially indistinguishable, all the possible combinations should be studied for a full statistical analysis.

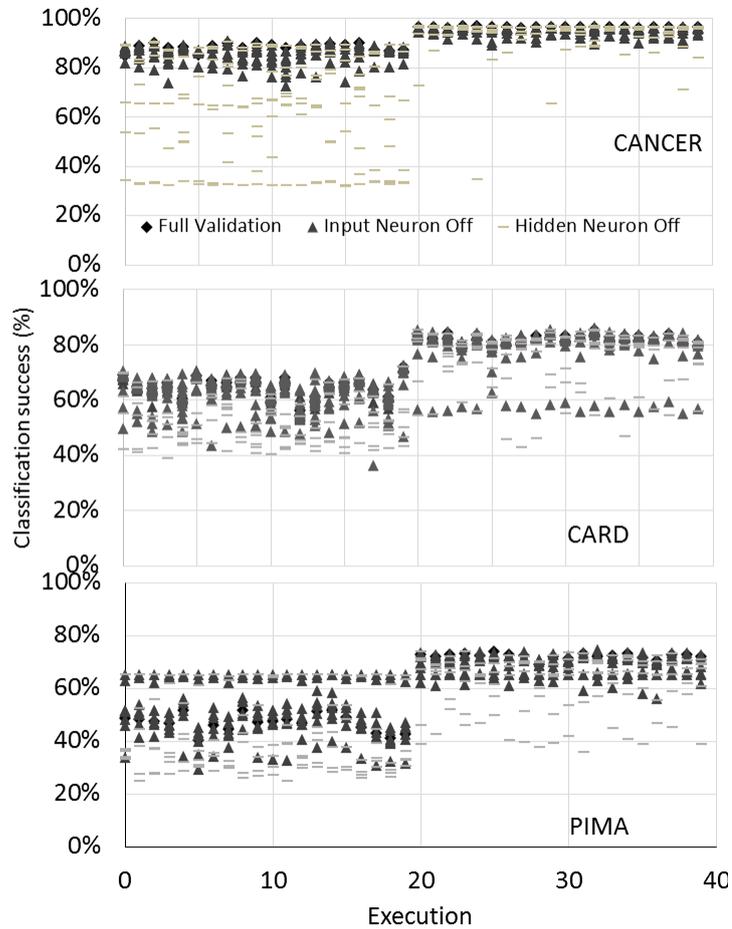

Figure 4: Classification success (defined as the number of cases that are correctly classified) for CANCER, CARD and PIMA sets. The figure shows simultaneously the values for 20 executions for both Backpropagation training (from 0 to 19) and Genetic Algorithm training (from 20 to 39). In all figures, we have a data set for every hidden and input neuron that is included in the FFNN geometry selected for every set under study.

Results of turning off a single hidden or input neuron are shown in figures 4 and 5. As we expected, BP trained networks are in general affected by the loss of many hidden neurons. Indeed, there are others that induce a much smaller effect when stopping their activity. This different influence of the hidden neurons has been extensively studied, for example, in several articles related to the pruning effect. Much more remarkable is the extremely high tolerance that is observed in the case of EA trained networks. In this case, the network is still having very similar ratios no matter which neuron is turned off. In table 2 and 3 we show the average and standard deviation respectively for all the cases under study. These values have been obtained by averaging the values from the suppression of every neuron in the 20 cases under study. As we can see, in all the cases we obtain better average values and smaller standard deviations in the EA trainings. The higher robustness against a non-working neuron makes EA training a better choice for the simulation of real biological systems. These kinds of systems must be stable against many effects, like the death of a certain number of neurons, which could imply the loss of the signal from any unit. Although biological systems have many different ways to prevent wrong network behaviour when the individual units are not working properly, an adequate weight distribution could be also an effective method to prevent the loss of information. In that sense, EA training have demonstrated to be a method with high capability of keep working practically in the same effective way when neurons are not having a perfect performance.

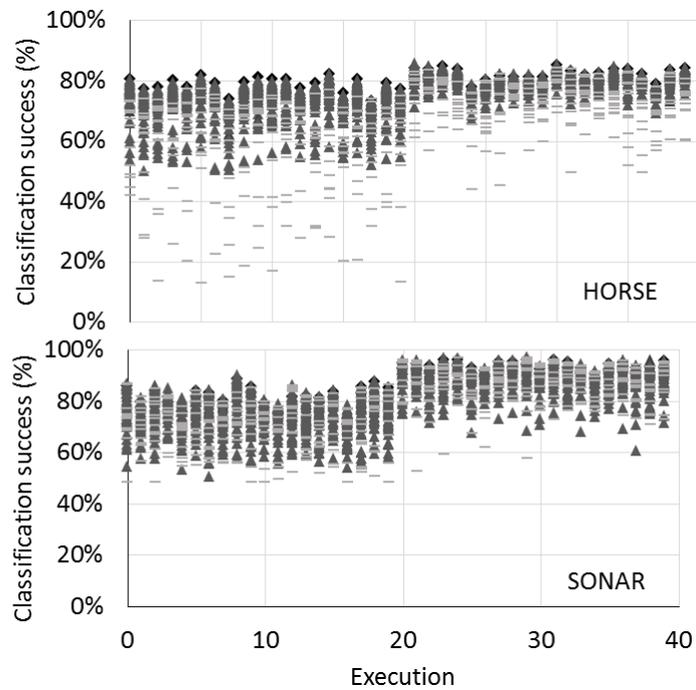

Figure 5: Classification success (defined as the number of cases that are correctly classified) for HORSE and SONAR sets. The figure shows simultaneously the values for 20 executions for both Backpropagation training (from 0 to 19) and Genetic Algorithm training (from 20 to 39). In all figures, we have a data set for every hidden and input neuron that is included in the FFNN geometry selected for every set under study.

There are a few effects in the figures and tables that must be mentioned in the analysis. First, the PIMA set seems to give better results when a hidden or input neuron is switched off. In figure 4 we can see the original set having values around 50%, which is approximately the value obtained from table 2. However, in many cases, we can see that this value can be increased up to 65% when neurons are switched off after the training. This effect is not really an improvement since this set has 503/771 inputs that correspond to a certain class. As we show in table 1, this numbers correspond to a relative value of 65.24%, which correspond to the maximum classification success percentage achieved in figure 4. What is really happening is that, sometimes, the modification included in the system makes it fail and the FFNN always answers with the second class, making it increase the classification success to 65.24% accidentally.

|        | Original |       | Hidden |       | Input |       |
|--------|----------|-------|--------|-------|-------|-------|
| Set:   | BP       | EA    | BP     | EA    | BP    | EA    |
| CANCER | 88.35    | 96.52 | 62.18  | 94.27 | 85.97 | 95.3  |
| CARD   | 65.29    | 82.86 | 52.54  | 75.67 | 64.67 | 82.17 |
| PIMA   | 48.09    | 72.22 | 41.58  | 62.55 | 50.9  | 68.82 |
| HORSE  | 78.89    | 82.57 | 58.26  | 72.67 | 73.3  | 80.59 |
| SONAR  | 82.45    | 94.55 | 70.45  | 86.94 | 76.63 | 90.75 |

Table 2: Average values of the classification success (%) for all the cases under study and both BP and EA training. Results are shown for the case where no modifications are given (original) and the cases where input and hidden neurons have been turned off.

|        | Original |      | Hidden |       | Input |      |
|--------|----------|------|--------|-------|-------|------|
| Set:   | BP       | EA   | BP     | EA    | BP    | EA   |
| CANCER | 1.51     | 0.19 | 21.58  | 6.44  | 3.58  | 1.6  |
| CARD   | 3.69     | 1.18 | 8.57   | 10.07 | 4.6   | 3.83 |
| PIMA   | 3.35     | 1.06 | 13.96  | 10.12 | 9.59  | 3.77 |
| HORSE  | 2.37     | 1.94 | 15.03  | 7.43  | 5.53  | 2.66 |
| SONAR  | 4.02     | 1.53 | 9.58   | 7.53  | 7     | 5.2  |

Table 3: Standard deviation (%) values of the classification success for all the cases under study and both BP and EA training. Results are shown for the case where no modifications are given (original) and the cases where input and hidden neurons have been turned off.

From figures 4 and 5, and tables 2 and 3 we can see that both training methods are much more stable when we switch off an input neuron. In figure 6 we show the average and standard deviation obtained for both BP and EA. We can see that losing an input neuron induces a minimum effect in the effectiveness of the FFNNs for both BP and EA. Interestingly, losing a hidden neuron does not have either a significant effect in the average value for EA in the CANCER set. Moreover, the effect is also smaller for the other 4 sets than in the case of BP. Analysing the SD, we see, however, that it has a significant increase even when the average seems to be similar. We can see the reason of this effect in figures 4 and 5, where some losing neurons reduce significantly the classification success for EA training.

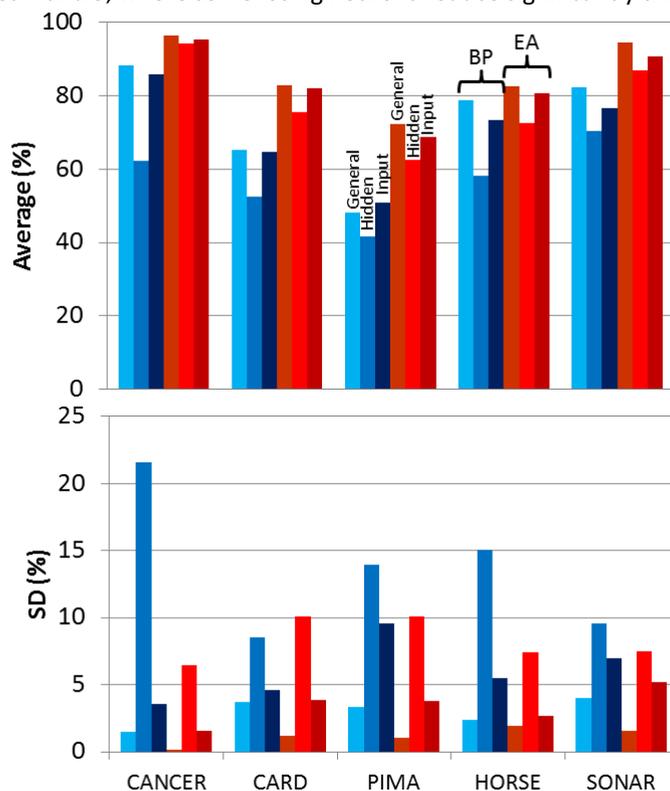

Figure 6: Average value and standard deviation of the classification success (%) for the five sets under study in the case of turning off one hidden or input neuron. The statistical calculations include the 20 different executions and all the hidden and input neuron modifications in the columns that represent the modification of the neurons in those layers.

Let us now make the problem worse by switching off two neurons from the hidden layer at the same time. In figures 7 and 8 we show the performance of the same sets than in figures 4 and 5 with the only exception that, in this case, we switch off all the possible combinations of two hidden neurons. Since the sets used in this article have very different number of input neurons, we did not include the effect of turning them off for clarity.

As expected, the performance is worse than in figures 4 and 5 for both BP and EA training. However, we can see that EA training is still able to give better results. In this case, it is not so clear to see the effect in the figures, due to the big amount of information. In this case, we need the help of the absolute values of average and standard deviation from tables 4 and 5.

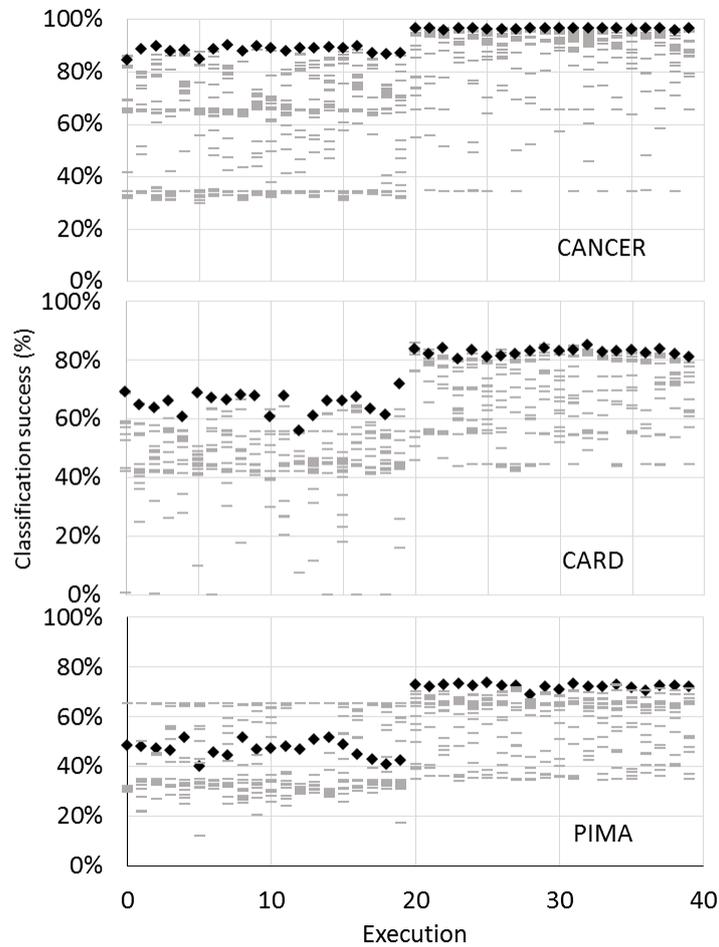

Figure 7: Classification success (defined as the number of cases that are correctly classified) for CANCER, CARD and PIMA sets. The figure shows simultaneously the values for 20 executions for both Backpropagation training (from 0 to 19) and Genetic Algorithm training (from 20 to 39). In all figures, we have a data set for every combination of two hidden neurons included in the FFNN geometry selected for every set under study.

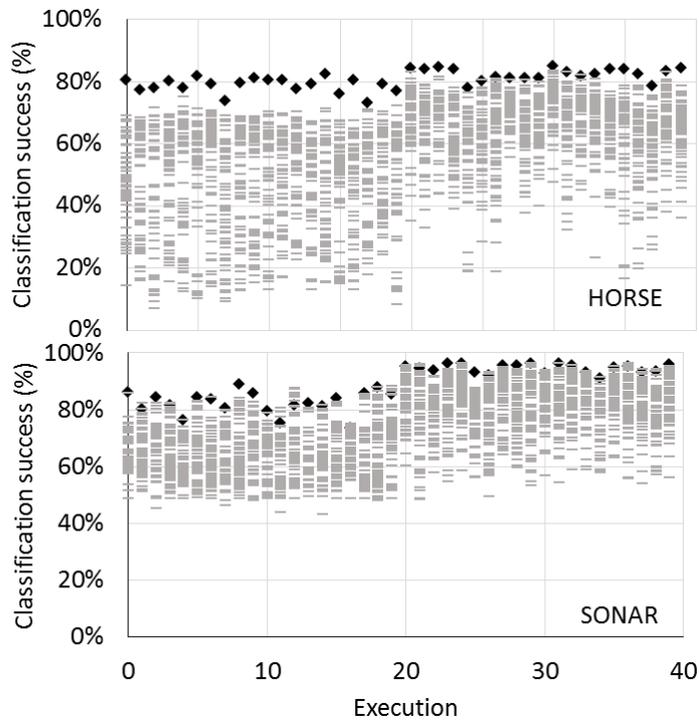

Figure 8: Classification success (defined as the number of cases that are correctly classified) for HORSE and SONAR sets. The figure shows simultaneously the values for 20 executions for both Backpropagation training (from 0 to 19) and Genetic Algorithm training (from 20 to 39). In all figures, we have a data set for every combination of two hidden neurons included in the FFNN geometry selected for every set under study.

From tables 4 and 5 we can see that switching off two neurons from the hidden layer in the case of EA training still gives better average results than the original case of BP in the CANCER, CARD and PIMA sets (being pretty similar in the SONAR set and clearly worse in HORSE). However, we must be very careful when comparing these results since removing a couple of hidden neurons always increases the standard deviation dramatically. This implies that, even when the average results are good, we cannot completely trust EA training now since there is a huge variation in the results depending on random factors such as the initialization of the weights. In figure 9 we show an image with the results from tables 4 and 5. In this figure we can see the small difference in the average value and, at the same time, the big increasing of the standard deviation.

|  | Original | | Hidden | |
| --- | --- | --- | --- | --- |
| Set: | BP | EA | BP | EA |
| CANCER | 88.35 | 96.52 | 55.71 | 89.69 |
| CARD | 65.29 | 82.86 | 46.33 | 68.86 |
| PIMA | 48.09 | 72.22 | 43.95 | 57.79 |
| HORSE | 78.89 | 82.57 | 48.22 | 64.76 |
| SONAR | 82.45 | 94.55 | 63.99 | 81.57 |

Table 4: Average values of the classification success (%) for all the cases under study and both BP and EA training. Results are shown for the case where no modifications are given (original) and the case where two hidden neurons have been turned off.

|  | Original | | Hidden | |
| --- | --- | --- | --- | --- |
| Set: | BP | EA | BP | EA |
| CANCER | 1.51 | 0.19 | 19.38 | 12.67 |
| CARD | 3.69 | 1.18 | 11.69 | 12.34 |
| PIMA | 3.35 | 1.06 | 15.25 | 11.63 |
| HORSE | 2.37 | 1.94 | 16.22 | 11.03 |
| SONAR | 4.02 | 1.53 | 9.54 | 9.41 |

Table 5: Standard deviation values (%) of the classification success for all the cases under study and both BP and EA training. Results are shown for the case where no modifications are given (original) and the case where two hidden neurons have been turned off.

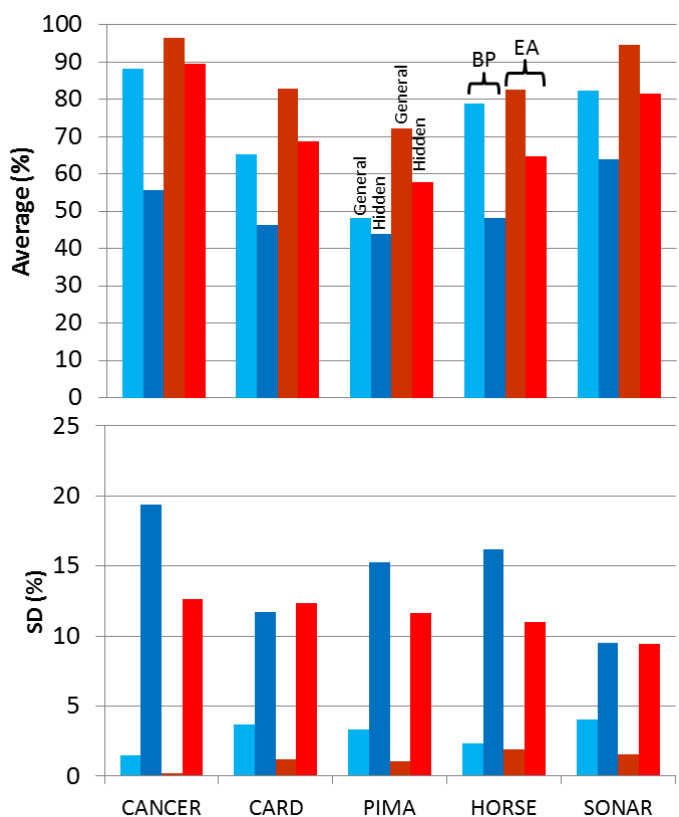

Figure 9: Average value and standard deviation of the classification success (%) for the five sets under study in the case of turning off two hidden neurons. The statistical calculations include the 20 different executions and all the hidden and input neuron modifications in the columns that represent the modification of the neurons in those layers.

### 4. Conclusions

We have shown a complete analysis of the effects of losing the neuron activity in feedforward neural networks. We demonstrate that there is a strong dependence between the performance of the network with the training method that has been used.

We have developed a method that allows us to give a different treatment to any neuron inside feedforward neural networks. The algorithm has been implemented inside two very different learning methods. The first one is a standard BP algorithm. The second one is an EA based training method, that has been fully described in the article. First, we have demonstrated that our EA training method converges faster and give more accurate results than BP. Then we have made a full analysis of the effects of turning off different combinations of neurons after the training phase is over.

EA training has demonstrated to be much more robust than BP for all the cases under study. Even in the strongest modification, when two hidden neurons are lost, EA training is able to give good average results. This difference implies that we must be very careful when pruning or redundancy effects are under study since the network performance when losing neurons strongly depends on the training method. Studies about the influence of the inputs are going to be also dependent on the training strategy.

From a different point of view, we have found a very robust training method when neurons are turned off (EA training). Using this method could be a good way to simulate biological learning systems since they must keep having a good performance against a deficient neuron performance. Although biological systems are much more complex than the simulations shown in this article, a smart training strategy such as the one shown here could be considered as a first defence against the losing of a certain number of neurons.

**ACKNOWLEDGEMENT**

Authors thank P. Varona, P. Molins-Ruano, C. González-Sacristán and F. Rodríguez for fruitful discussions. The authors acknowledge support from TIN-2010-19607. GMS acknowledges support from the Spanish Ramón y Cajal program.